\documentclass[conference]{IEEEtran}
\IEEEoverridecommandlockouts
\usepackage{times}
\usepackage{graphicx}
\usepackage{latexsym}
\usepackage{color}
\usepackage{amsmath}
\usepackage{amssymb}
\usepackage{amsthm}
\usepackage{multirow}
\usepackage[bookmarks=false]{hyperref}
\usepackage{tikz}
\usetikzlibrary{graphs,arrows,positioning,trees,decorations.pathmorphing,chains,shapes,decorations.pathreplacing,quotes,calc}
\def\BibTeX{{\rm B\kern-.05em{\sc i\kern-.025em b}\kern-.08em    T\kern-.1667em\lower.7ex\hbox{E}\kern-.125emX}}

\newcommand{\Tensor}[1]{\ensuremath{\mathbf{ #1}}}

\tikzstyle{hidden} = [circle,draw]
\tikzstyle{visible} = [circle,draw,fill=black!10]
\tikzstyle{myLink} = [->,>=latex]

\graphicspath{{plots/}}
\begin{document}

\title{Generalising Recursive Neural Models by Tensor Decomposition} %Proposta Davide
\author{\IEEEauthorblockN{Daniele Castellana and Davide Bacciu}
\IEEEauthorblockA{Dipartimento di Informatica\\
Universit\`a di Pisa\\
Largo B. Pontecorvo 3\\
Pisa, Italy\\
Email: \{daniele.castellana,bacciu\}@di.unipi.it}
}

    \maketitle

\begin{abstract}
    Most machine learning models for structured data encode the structural knowledge of a node by leveraging simple aggregation functions (in neural models, typically a weighted sum) of the information in the node's neighbourhood. Nevertheless, the choice of simple context aggregation functions, such as the sum, can be widely sub-optimal. In this work we introduce a general approach to model aggregation of structural context leveraging a tensor-based formulation. We show how the exponential growth in the size of the parameter space can be controlled through an approximation based on the Tucker tensor decomposition. This approximation allows limiting the parameters space size, decoupling it from its strict relation with the size of the hidden encoding space. By this means, we can effectively regulate the trade-off between expressivity of the encoding, controlled by the hidden size, computational complexity and model generalisation, influenced by parameterisation. Finally, we introduce a new Tensorial Tree-LSTM derived as an instance of our framework and we use it to experimentally assess our working hypotheses on tree classification scenarios.
\end{abstract}

\section{Introduction}
    Trees are complex data which arise in multiple contexts: in Natural Language Processing, parse trees are used to represent natural language sentences; on the web, most of the data (e.g. HTML and XML documents) are represented using the Document Object Model i.e. a tree which represents the document structure. Regardless of the application domain, trees represent hierarchical information: they are composed of atomic entities, called \emph{nodes}, combined together through a \emph{parent-child} relationship which defines the tree structure. 
    
    The first approaches for learning from tree data were developed in the nineties; models, such as the \emph{RAAM} \cite{Pollack1990} and the \emph{LRAAM} \cite{Sperduti94labelingraam}, and theories, such as the \emph{Backpropagation Through Structure} (BPTS) \cite{Goller1996} and the \emph{generalised recursive neuron} \cite{Sperduti1997}, were elaborated in those years. All these concepts have been unified in a general framework for processing of structured data \cite{Frasconi1998} through, so-called, recursive learning models.%, which describes how to develop a machine learning model to tackle supervised learning problems on tree-structured data.
    
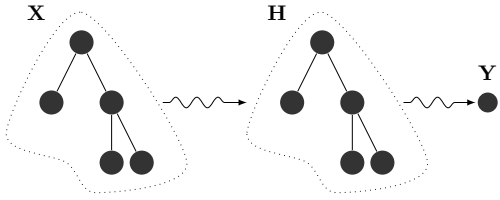
\begin{figure}
        \centering
        \tikzstyle{filled} = [circle,fill=black!80]
        \scalebox{0.8}{
        \begin{tikzpicture}[decoration = {snake, pre length=3pt,post length=7pt}]
	        \node (x) at (-2,0) {
                \begin{tikzpicture}[edge from parent/.style={draw,-}, every node/.style = {minimum size = 0.4cm}, level 1/.style={sibling distance=0.5cm, level distance=1cm} ,, node distance = 0.1cm]
                    \node [filled] (q1) {}
                    child {node [filled] (q2) {}}
                    child [missing]
                    child {node [filled] (q3) {}
                        child [missing]
                        child {node [filled] (q4) {}}
                        child {node [filled] (q5) {}}
                    };
                \end{tikzpicture}
                };
        	\node (h) at (2,0) {
                \begin{tikzpicture}[edge from parent/.style={draw,-}, every node/.style = {minimum size = 0.4cm}, level 1/.style={sibling distance=0.5cm, level distance=1cm} ,node distance = 0.1cm]
                    \node [filled] (q1) {}
                    child {node [filled] (q2) {}}
                    child [missing]
                    child {node [filled] (q3) {}
                        child [missing]
                        child {node [filled] (q4) {}}
                        child {node [filled] (q5) {}}
                        };
                \end{tikzpicture}
                };
        	\node [filled] (y) at (4.5,0) {};
        	\draw [decorate,myLink] ($(x) + (1.1,0)$)-> ($(h) + (-1.5,0)$) node [midway, yshift =2em ] (T) {};%{$\hat{f}$};
        	\draw [decorate,myLink] ($(h) + (1.1,0)$) -> ($(y) + (-0.2,0)$) node [midway, yshift =2em ] (T) {};%{$\tau$};
        	\draw [dotted] plot[smooth cycle, tension=0.5] coordinates {(-3.5,-0.5) (-2.25,1.5) (-0.5,-1) (-2,-1.5) (-2.5,-1)};
        	\draw [dotted] plot[smooth cycle, tension=0.5] coordinates {(1.5,-1) (0.5,-0.5) (1.75,1.5) (3.5,-1) (2,-1.5)};
        	\node at (1,1.5) {$\mathbf{H}$};
        	\node at (-3,1.5) {$\mathbf{X}$};
        	\node at (4.5,0.5) {$\mathbf{Y}$};
        \end{tikzpicture}
        }
        \caption{An example of a tree sample $\mathbf{X}$, its hidden representation $\mathbf{H}$ and corresponding output label $\mathbf{Y}$.}
        \label{fig:tree_problem_example}
    \end{figure}

    In this paper we focus on supervised tasks which admit a \emph{recursive state representation}, i.e. it exists a hidden representation $\mathbf{H}$ which makes the output $\mathbf{Y}$ independent from the input $\mathbf{X}$ \cite{Frasconi1998}. The hidden state $\mathbf{H}$, the input $\mathbf{X}$ and the output $\mathbf{Y}$ are trees with the same skeleton (classification tasks can be implemented as special case through a deterministic reduction of structure $\mathbf{Y}$  as in \cite{Frasconi1998}, see Figure \ref{fig:tree_problem_example}). Moreover, the hidden representation $\mathbf{H}$ can be computed recursively on the structure. The latter property induces the concept of \emph{context} of a node, i.e. the minimal set of nodes which contains all the necessary information to compute the target node hidden state. Clearly, the definition of a context is strictly related to the partial order used to define the recursion. For example, if the hidden state is computed recursively top-down, it means that the context of each tree node is its parent node; the root, which has no parent, is the base case of the recursion. Vice versa, if the recursion direction is bottom-up, the context of a node is the set of all its children; leaf nodes, which have no children, are the recursion base case.
    
    The effect of the parsing direction on the properties and complexity of the model has been widely studied in the context of generative models \cite{bacciu2018learning,Bacciu2012}. On the other hand, neural models tend to be less studied from this perspective. In particular, bottom-up neural models seem immune to the exponential growth of the parameters space with respect to the tree output degree which instead is considered a limiting factor in bottom-up generative models where approximations are required \cite{Bacciu2012, Bacciu2013CompositionalModel, Castellana2019}.
    
    Within the scope of this paper, we attempt to fill this gap by showing that there exists a function which combines children hidden states through a tensor operator also in the case of $L$-ary trees and that this leads to an exponential number of learnable parameters. Moreover, we discuss how a formulation of the recursive neural state transition function in terms of such tensor operator is a generalisation of the weighted summation that is usually adopted in the literature for the same purpose.% Unfortunately, the parameters space explosion makes this approach infeasible in practice.
    
    A central contribution of this paper is an approximation of such tensor state transition function which leverages a tensor decomposition known as Tucker decomposition \cite{Tucker1966}. This approximation introduces a new hyper-parameter which controls the complexity of the parameters space of the transition function, breaking up the exponential dependence on hidden state size. This allows combining together the advantages of sum-based and tensor-based state transitions. In particular, we enable to deploy a model with a large hidden space (as it happens in models using simple sum-based aggregation) while computing state transitions trough a powerful aggregation function (as it happens in the full tensor model). Note that larger hidden states increase the ability of the neural model of representing diverse and articulated substructures by means of the activation of its hidden neurons. On the other hand, the use of tensor-based aggregation allows the state transition function to explicitly capture direct dependencies between the encodings of the children of a node.
    
    As part of this work, we also introduce two new Tree-LSTM based models using, respectively, the full tensor state transition function and its Tucker-based approximation. While tensor decompositions are already used in neural networks to compress full neural layers, this is the first work that, to the extent of our knowledge, leverages tensor decomposition as a more expressive alternative aggregation function for neurons in structured data processing. Among the others, we experimentally assess the advantages that this yields in terms of decoupling the parameter space size and the hidden state size, testing our models and a standard sum-based Tree-LSTM \cite{shortedTai2015} on two challenging tree classification benchmarks. 
        
\section{Related Work}
    Using a tensor to compose input information has been applied successfully in other contexts: for example, in  \cite{ben2017mutan}, the authors use tensor to fuse visual and textual representations; in \cite{Weber2017EventRW}, the authors use tensors to combine predicate, object and subject in event triplets. In both papers, the number of input to combine is fixed to two or three. Our proposal differs from these works because we use tensors to compose several inputs whose number is not fixed but depends on input data.
    
    Regarding the processing of structured data, the first high-order neural network to process tree-structured contextual information has been discussed in \cite{Frasconi1998}. However, the authors provided only the formulation for the case of binary trees, without any discussion on its properties or any experimental analysis. As far as we know, the only implementation of an high-order neural network for structured data is the one discussed in \cite{shortedSocher2013}. Again, only the case of binary trees is considered. Our framework extends this works taking into account $L$-ary trees and by introducing the use of tensor decompositions as a viable and expressive tradeoff between the simplicity of sum aggregation and the complexity of full tensor aggregations.
    %where $L$ is the maximum number of child nodes.
    
    Tensor decompositions also aroused much interest in the community thanks to their compression ability. For example, in \cite{DBLP:Calvi, NoikovTensorizing}, the authors show their use in compressing a fully connected layer with minor reduction in performance. The compression is obtained straightforwardly by storing the weight matrix using tensor decomposition. Differently, in this work, we propose tensor decompositions as an alternative to weighted sum input aggregation in  recursive neurons, focusing in particular on the expressiveness of the newly proposed aggregation function.
    
\section{Neural Tensor State Transition Function}\label{sec:staete_transition} 
    The \emph{state transition} function, in structured data processing, is the recursive function which computes the hidden representation of a node $v$ given its label $x_v$ and its associated context, e.g. the ordered set of its children $v_1, \dots, v_L$. In neural terms, the state transition function can be interpreted as the encoding of $v$ corresponding to the vector of hidden neuron activations for the node.  Without loss of generality, let us focus on a bottom-up state transition function $\hat{f}$, defined as \cite{Frasconi1998}:
    \begin{equation}
        h_v = \hat{f}\left(x_v,h_{v1}, \dots. h_{vL}\right),
    \end{equation} 
    where $h_v$, $x_v$ and $\{h_{v1}, \dots, h_{vL},\}$ are the hidden state, the input label and the context of target node $v$, respectively. The context contains $L$ hidden states, where $L$ is the maximum outdegree, i.e. the maximum number of child nodes. The term $h_{vj}$ denotes the hidden state of the $j$-th child node of $v$.
    
    The function $\hat{f}$ models the relationships between the pieces composing the structured sample (i.e. the nodes); $\hat{f}$ is typically unknown and, hence, the goal is estimating it from data. In neural models, $\hat{f}$ is typically approximated through the composition of a non-linear activation function $\sigma$ and an affine map $f$:
    \begin{equation}
        h_v =  \sigma \left( f\left(x_v,h_{v1}, \dots, h_{vL}\right) \right).
        \label{eq:generalRNN}
    \end{equation}
    When dealing with flat (vectorial) data, $f$ takes only one input (the label $x_v$ since there is no context). In this case, the affine map is defined as $f(x_v)=Wx_v + b$ where $W$ is a matrix and $b$ a vector; equivalently, using the homogeneous coordinate $\bar{x}_v = [x_v; 1]$,  $f(x_v) = \bar{W}\bar{x}_v$  where $\bar{W}$ is the \emph{augmented matrix} obtained concatenating $b$ to the last column of $W$. The operator $[a;n]$ adds the scalar $n$ as the last element of the vector $a$.
    
    With tree data, instead, we also consider the contribution of a context of size $L$. Therefore, the function 
    \begin{equation} \label{eq:affinef}
    f: \mathbb{R}^m \times  \overbrace{\mathbb{R}^c \times \dots \times \mathbb{R}^c}^{L \text{ times}} \rightarrow \mathbb{R}^c
    \end{equation}
    becomes a multi-affine map, where $m$ and $c$ denote the size of the input label and hidden states, respectively (i.e. $x_v \in \mathbb{R}^m$ and $h_v \in \mathbb{R}^c$).
    
    In this work, we assume that the multi-affine map $f$ is represented as an \emph{augmented tensor} $\Tensor{T} \in \mathbb{R}^{(m+1) \times (c+1) \dots \times (c+1) \times c}$ (in the supplementary material\footnote{\url{http://pages.di.unipi.it/castellana/publication/generalise_tree_rnn/}} we prove that such tensor represents a multi-affine map). The last dimension of the tensor is associated to the output vector. As for the matrix case, each input vector $v$ is represented using the homogeneous coordinate $\bar{v} = [v; 1]$. %augmented tensor allows us to use homogeneous coordinates to represent each input vector and therefore to represent affine transformations.
    
    In the following, we use $f^\Tensor{T}$ to denote that $f$ is a multi-affine map defined by the tensor \Tensor{T}. Therefore, we can rewrite Eq. \eqref{eq:tensorRNN} as:
    \begin{multline}
        h_v(k) = \sigma(f^\Tensor{T}(x_v,h_{v1},\dots,h_{v_L}))(k) =\\
        =\sigma\left(\sum_{i=1}^{m+1} \sum_{j_1=1}^{c+1}\dots\sum_{j_L=1}^{c+1}\Tensor{T}(i,j_1, \dots,j_L, k) \times \right.\\
        \times \left. \bar{x}(i)\bar{h}_1(j_1)\dots \bar{h}_L(j_L)  \vphantom{\sum_{i=1}^{m+1}}\right).
        \label{eq:tensorRNN}
    \end{multline}
    We use brackets to denote entries of vectors, matrices and tensors: for instance, $W(i,j)$ denotes the element of the matrix $W$ in the $i$-th row and $j$-th column. 
    
    Each entry of the tensor \Tensor{T} should be learned during the training phase to approximate the true state transition function $\hat{f}$. Hence, we have a number of parameters which is equal to $(m+1)(c+1)^L c \approx O(mc^{(L+1)})$.
    
    The exponential growth of parameters space with respect to outdegree $L$ imposes strong limitations on the choice of the hidden state size (i.e. the number of hidden neurons in an internal layer). With a full tensorial formulation, even for small values of $L$ (e.g. $L=5$), a hidden state of size $10$ would require about one million of parameters. This limitation can lead to poor performances since a small hidden state can be insufficient to encode all structural relationships and topological variations that a rich representation such as a tree can bring.
    %The equation \eqref{eq:tensorRNN} does not make any independence assumption among the input label and child hidden states. The price to pay for such expressive power is the exponential growth of the tensor $\Tensor{T}$ size w.r.t. to the maximum output degree $L$. 
    For this reason, the prominent approach is to avoid the expressive but exponential formulation in Eq. (\ref{eq:tensorRNN}) in favour of a simplified hidden state transition function which allows deploying a relatively large number of hidden neurons. In Section \ref{sec:sum_approx}, we formally show how such widely used transition function can be derived as an approximation of our general formulation in Eq. \eqref{eq:tensorRNN}. In Section \ref{sec:hosvd_appox}, instead, we introduce a new approximated state transition function which builds on the Tucker tensor decomposition. This decomposition introduces a new hyper-parameter which breaks the exponential relation between the parameters space size and hidden state size. By this means, it allows deploying a large hidden state size without incurring in an exponential growth of the number of model parameters.

    \subsection{Weighted Sum Approximation} \label{sec:sum_approx}
    In most neural models for trees \cite{Frasconi1998, shortedSocher2013}, the state transition function is defined as:
    \begin{equation}
        h_v = \sigma\left(W x_v + \sum_{j=1}^L U_j h_{vj} + b\right).
        \label{eq:sumRNN}
    \end{equation}
    The hidden state $h_v$ is obtained by applying a linear transformation $W$ to the input label $x_v$ and a linear transformation $U_j$ to each child hidden state $h_{vj}$. The results of such transformations are summed together with a bias $b$, and fed to a non-linear function $\sigma(\cdot)$. The summation defines a multi-affine map $f^+$ which can be rewritten as:
    \[
        f^+(x,h_1,\dots,h_L) = Wx + U_1h_1 + \dots + U_Lh_L + b.
    \]
    
    In this formulation, each input vector (from label or context) contributes separately to the computation of the hidden state $h_v$ and, therefore, it is completely independent from the others. This section aims to prove that $f^+$ can be obtained by imposing a full independence assumption among input vectors of the multi-affine map $f^\Tensor{T}$ in Eq. \eqref{eq:tensorRNN}. To this end, we show there exists a tensor $\Tensor{T}^+$ such that:
    \begin{multline}
        f^+(x,h_1,\dots,h_L)(k) =f^{\Tensor{T}^+}(x,h_1,\dots,h_L)(k)\\
        =\sum_{i=1}^{m+1} \sum_{j_1=1}^{c+1}\dots\sum_{j_L=1}^{c+1}\Tensor{T}^+(i,j_1, \dots,j_L, k) \times \\
        \times \bar{x}(i)\bar{h}_1(j_1)\dots \bar{h}_L(j_L).
        \label{eq:tensor_sum}
    \end{multline}
    For the sake of clarity, we have omitted the subscript related to current node $v$. In order to impose independence between input vectors, we must have all those entries of $\Tensor{T}^+$ which are multiplied with more than one input element set to $0$. In fact, in $f^+$ above, there are no terms with inputs multiplied together. For example, the entry $\Tensor{T}^+(1,1,\dots,1, k)$ is multiplied by the first element of each input vector (i.e. $x(1), h_1(1), \dots, h_L(1)$) and therefore must be equal to $0$. 
    
    Recalling that we are using homogeneous coordinates (hence the last element of each input vector is equal to $1$), the only non-zero entries in $\Tensor{T}^+$ are the ones which have all input indexes (except one) pointing to the last entry of the input vector. For example, the entry $\Tensor{T}^+(i,c+1, \dots,c+1, k)$ multiplies solely to $x(i)$, since $\bar{h}_1(c+1) = \dots = \bar{h}_L(c+1) = 1$. Therefore, we obtain the following equation:
    \begin{multline}
        h(k) = f^+(x,h_1,\dots,h_L)(k) =\\
        =\sum_{i=1}^{m} \Tensor{T}^+(i,c+1, \dots,c+1, k)x(i) + \\
        + \sum_{j_1=1}^{c}\Tensor{T}^+(m+1,j_1, \dots,c+1, k)h_1(j_1) + \\
         \dots \\
        + \sum_{j_L=1}^{c}\Tensor{T}^+(m+1,c_1, \dots,j_L, k)h_L(j_L) + \\
        + \Tensor{T}^+(m+1,c+1, \dots,c+1, k),
        \label{eq:tensor_sum2}
    \end{multline}
    where each row in the equation, except the last one, contains the application of a linear map; the last row contains a vector. Finally, Eq. \eqref{eq:tensorRNN} can be derived by setting each entry of the tensor to:
    \begin{multline}
        \Tensor{T}^+(i, j_1,\dots,j_L,k) = \\
        =\begin{cases}
        W(i,k) & \text{if } j_1=\dots=j_L = c+1\\
        U_l(j_l,k) & \text{if } i=m+1 \land j_{\neq l}=c+1\\
        b(k) & \text{if } i=m+1 \land j_1=\dots=j_L = c+1\\
        0 & \text{otherwise}
        \end{cases}.
    \end{multline}
    This approximation requires $cm + L c^2 + c \approx O(Lc^2)$ parameters, removing the exponential dependence between the number of parameters and the hidden state size.
    
    \subsection{HOSVD Approximation} \label{sec:hosvd_appox}
    The general tensor \Tensor{T} in Eq. \eqref{eq:tensorRNN} can be approximated using a tensor factorisation. To this end, we propose to use the factorisation known as \textit{Tucker} decomposition; it was originally developed by Tucker to obtain a method for identifying relations in a three-way tensor of psychometric data \cite{Tucker1966}. Later, the same idea has been generalised to a $p$-way tensor proving that it is a multi-linear generalisation of the matrix Singular Value Decomposition (SVD) \cite{DeLathauwer2000}. Hence, the same decomposition is also called High-Order Singular Value Decomposition (HOSVD) \cite{DeLathauwer2000}.
    
    Let $\Tensor{T} \in \mathbb{R}^{I_1 \times \dots \times I_p}$ be a tensor with $p$ dimensions, each entry $\Tensor{T}(i_1,\dots, i_p)$ can be factorised as \cite{DeLathauwer2000}:
    \begin{multline}
    \Tensor{T}(i_1,\dots, i_p) =\\
    =\sum_{j_1=1}^{R_1} \cdots \sum_{j_p=1}^{R_p} \Tensor{G}(j_1,\dots,j_p) \prod_{z=1}^p U_{z}(j_z,i_z),
    \label{eq:Tucker_decomp}
    \end{multline}
    where $\Tensor{G}(j_1,\dots,j_p)$ are elements of a new $p$-way tensor $\Tensor{G}\in \mathbb{R}^{R_1 \times \dots \times R_p}$ called \emph{core} tensor and $U_z(j_z,i_z)$ are elements of matrices $U_z \in \mathbb{R}^{R_z \times I_z}$ called \emph{mode} matrices. The value $R_z$ is the \emph{rank} along the $z$-th dimension.% An example of decomposition can be visualised in Figure \ref{fig:Tucker_decomp}.

    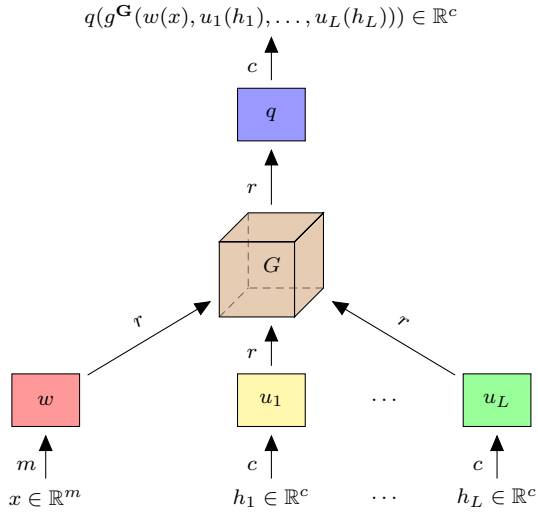
\begin{figure}[tb]
        \centering
        \begin{tikzpicture}[edge from parent/.style={draw,<-,>=triangle 45}, every node/.style = {minimum size = 0.5cm}, execute at begin node=$, execute at end node=$,node distance = 0.5cm,level 1/.style={level distance=1.3cm},level 2/.style={level distance=2cm}, level 3/.style={level distance=1.8cm},level 4/.style={level distance=1.3cm}]
            \footnotesize
			\pgfmathsetmacro{\cubex}{1}
			\pgfmathsetmacro{\cubey}{1}
			\pgfmathsetmacro{\cubez}{1}
			\pgfmathsetmacro{\corex}{0.9}
			\pgfmathsetmacro{\corey}{0.7}
			\node [minimum size=0cm] (qu) {q(g^{\Tensor{G}}(w(x),u_1(h_{1}),\dots,u_L(h_{L}))) \in \mathbb{R}^c}
			child { 
					node[] (q) {
						\begin{tikzpicture}[every edge quotes/.append style={auto, text=blue}]
							\draw [draw=black, every edge/.append style={draw=black, densely dashed, opacity=.5}, fill=blue!40] (0,0,0) coordinate (o) -- ++(-\corex,0,0) coordinate (a) -- ++(0,-\corey,0) coordinate (b) -- ++(\corex,0,0) coordinate (c) -- cycle;
							\node at (-0.45,-0.35) {q};
						\end{tikzpicture}
					}
    				child {  
						node (G){
							\begin{tikzpicture}[every edge quotes/.append style={auto, text=blue}]
								\draw [draw=black, every edge/.append style={draw=black, densely dashed, opacity=.5}, fill=brown!40]
								(0,0,0) coordinate (o) -- ++(-\cubex,0,0) coordinate (a) -- ++(0,-\cubey,0) coordinate (b) edge coordinate [pos=1] (g) ++(0,0,-\cubez)  -- ++(\cubex,0,0) coordinate (c) -- cycle
								(o) -- ++(0,0,-\cubez) coordinate (d) -- ++(0,-\cubey,0) coordinate (e) edge (g) -- (c) -- cycle
								(o) -- (a) -- ++(0,0,-\cubez) coordinate (f) edge (g) -- (d) -- cycle;
								\node at (-0.3,-0.3) {G};
							\end{tikzpicture}
						} 
						child{
							%node (wx){w(x)}  
							%child{
								node [minimum size=0cm] (w) {
			    						\begin{tikzpicture}[every edge quotes/.append style={auto, text=blue}]
				    						\draw [draw=black, every edge/.append style={draw=black, densely dashed, opacity=.5}, fill=red!40]
				    						(0,0,0) coordinate (o) -- ++(-\corex,0,0) coordinate (a) -- ++(0,-\corey,0) coordinate (b) -- ++(\corex,0,0) coordinate (c) -- cycle;
				    						\node at (-0.45,-0.35) {w};
			    						\end{tikzpicture}
		    						} 
		    						child{ node {x \in \mathbb{R}^m} 
		    						    edge from parent node[pos=0.7,left,draw=none] {m} 
		    						}
							%}
							edge from parent node[sloped,above,draw=none] {r}  
						}
						child{
							node {} edge from parent[draw=none]
							child{
								node [minimum size=0cm] {} edge from parent[draw=none]
							}
						}
						child{
							%node (uh1){u_1(h_1)}
							%child{
								node [minimum size=0cm] (u_1) {
			    						\begin{tikzpicture}[every edge quotes/.append style={auto, text=blue}]
				    						\draw [draw=black, every edge/.append style={draw=black, densely dashed, opacity=.5}, fill=yellow!40]
				    						(0,0,0) coordinate (o) -- ++(-\corex,0,0) coordinate (a) -- ++(0,-\corey,0) coordinate (b) -- ++(\corex,0,0) coordinate (c) -- cycle;
				    						\node at (-0.45,-0.35) {u_1};
			    						\end{tikzpicture}
		    						}
		    						child{ node {h_1 \in \mathbb{R}^c} 
		    						    edge from parent node[pos=0.7,left,draw=none] {c}
		    						}
							%}
							edge from parent node[pos=0.7,left,draw=none] {r}  
						}
						child{
							node {\dots} edge from parent[draw=none]
							child{
								node [minimum size=0cm] {\dots} edge from parent[draw=none]
							}
						}
						child{
							%node (uhL){u_L(h_L)}
							%child{
								node [minimum size=0cm] (u_1) {
			    						\begin{tikzpicture}[every edge quotes/.append style={auto, text=blue}]
				    						\draw [draw=black, every edge/.append style={draw=black, densely dashed, opacity=.5}, fill=green!40]
				    						(0,0,0) coordinate (o) -- ++(-\corex,0,0) coordinate (a) -- ++(0,-\corey,0) coordinate (b) -- ++(\corex,0,0) coordinate (c) -- cycle;
				    						\node at (-0.45,-0.35) {u_L};
			    						\end{tikzpicture}
		    						}
		    						child{ node {h_L \in \mathbb{R}^c} 
		    						    edge from parent node[pos=0.7,left,draw=none] {c}
		    						}
							%}
							edge from parent node[sloped,above,draw=none] {r}  
						}
    				    edge from parent node[pos=0.7,left,draw=none] {r} 
				    }
    			edge from parent node[pos=0.7,left,draw=none] {c} 
    		};
        \end{tikzpicture}
        \caption{A graphical representation of the HOSVD state transition function. The input $(x,h_1, \dots, h_L)$ are transformed using the mode matrices $(w, u_1,\dots, u_L)$ respectively. Then, the core tensor \Tensor{G} is used to combine the transformed values; its output is fed to the last mode matrix $q$. Letters on links indicate the dimension of the vector sent along that direction.}
        \label{fig:hosvdRNN}
    \end{figure}
    
    Ranks play a fundamental role since the number of parameters needed by the approximation is strictly related to it. In fact, we need $\prod_{z=1}^p R_z$ parameters to represent the core tensor and $\sum_{z=1}^p R_z \times I_z$ to represent the mode matrices. Therefore, the lower the ranks, the more we can compress the original tensor \Tensor{T}. Even if we have the guarantee that a HOSVD exists for every tensor \cite{DeLathauwer2000}, we have no guarantee on the rank values. In the worst case, $R_z = I_z$ and the core tensor has the same dimension as the initial one; hence, there is no compression.

    Within the context of our neural functions, we would like to use the HOSVD to limit the parameters explosion of the general tensor state transition function. To this end, it is convenient to interpret the HOSVD in terms of mappings. Let $f^\Tensor{T}$ be the state transition function represented by a tensor \Tensor{T}, then we can decompose $f^\Tensor{T}$ by applying the HOSVD as
    \begin{equation}
         f^\Tensor{T}(x,h_1,\dots,h_L) \approx q(g^{\Tensor{G}}(w(x),u_1(h_{1}),\dots,u_L(h_{L}))),
         \label{eq:hosvdRNN}
    \end{equation}
    where $w(\cdot)$ is the linear map represented by the mode matrix $W$ along the first dimension (i.e. the dimension of the input label). The terms $u_1(\cdot), \dots, u_L(\cdot)$ denote the linear maps represented by the mode matrices $U_1,\dots,U_L$ along the other input dimensions (from the second one to to the second-to-last), while $g^\Tensor{G}$ is a multi-affine map represented by the core tensor \Tensor{G}. Finally, $q(\cdot)$ is the linear map represented by the mode matrix $Q$ along the output dimension (the last one). We report in Figure \ref{fig:hosvdRNN} a graphical representation of Equation \eqref{eq:hosvdRNN}.
    
    %Thanks to the core tensor, the HOSVD approximation allows the state transition to combine input information using a function which is more expressive than the weighted sum.
    Thanks to the core tensor, the HOSVD approximation allows the state transition to combine input information without assuming full independence on them (as in the weighted sum approximation). Nevertheless, such combination is performed in a reduced space cutting down the number of parameters required. The value of the rank $r$, that we assume equal for each dimension, indicates the size of the reduced input space. Hence, the size of the tensor ultimately performing the computations needed for the state transition depends on the rank rather than on the size of the hidden states. More in detail: the mode matrix $W$ requires $m\times r$ parameters; each of the mode matrices $U_1 \dots, U_L$ and $Q$ require $c\times r$ parameters; the core tensor \Tensor{G} requires $r(r+1)^{(L+1)}$ parameters. Hence, the HOSVD approximation requires a total of $mr + (L+1)cr + r(r+1)^{(L+1)}$ parameters.
    
    \iffalse
    Using the tensor notation, we can rewrite the equation \eqref{eq:hosvdRNN} as:
    \begin{multline}
        f(x,h_1,\dots,h_L)(k) = \sigma(g(x,h_{1},\dots,h_{L}))(k) =\\
        =\sigma\left( \sum_{q=1}^{R} Q(q,k)  \sum_{t=1}^{S} \sum_{z_1=1}^{R_1} \cdots \sum_{z_L=1}^{R_L}\Tensor{G}(t,z_1,\dots,z_L, k)  \times \right.\\
        \times \sum_{i=1}^{m} W(t,i)x(i) \sum_{j_1=1}^{c} U_1(z_1,j_1)h_1(j_1) \dots \\
        \dots \left. \sum_{j_L=1}^{c} U_L(z_L,j_L)h_L(j_L)  \vphantom{\sum_{i=1}^{m+1}}\right).
        \label{eq:hosvdRNN_tensor}
    \end{multline}
    \fi
    
\section{TENSOR TREE-LSTM MODELS}\label{sec:tree-LSTM}
    The Tree-LSTM \cite{shortedTai2015} is a generalisation of the standard LSTM architecture to tree-structured data. As in the standard LSTM, each LSTM unit contains: the \emph{input gate}, the \emph{output gate}, a \emph{forget gate} for each child node, and the \emph{memory cell}. For our purpose, we focus on the L-ary Tree-LSTM defined by the following equations \cite{shortedTai2015}:
    \begin{equation}
        \begin{split}
            i_v &= \sigma\left(W^i x_v + \sum_{j=1}^L U^i_j h_{vj} + b^i\right),\\
            o_v &= \sigma\left(W^o x_v + \sum_{j=1}^L U^o_j h_{vj} + b^o\right),\\
            u_v &= \sigma\left(W^u x_v + \sum_{j=1}^L U^u_j h_{vj} + b^u\right),\\
            f_{vj} &= \sigma\left(W^f x_v + U^f_{j} h_{vj} + b_j^f\right),\\
            c_v &= i_v \odot u_v + \sum_{j=1}^{L} f_{vj} \odot c_{vj},\\
            h_v &= o_v \odot \text{tanh}(c_v);
        \end{split}
        \label{eq:sumLSTM}
    \end{equation}
    where $i_v$, $o_v$, $c_v$ are the input gate, the output gate and the memory cell, respectively, and $f_{vj}$ is the forget gate associated to the $j$-th child node. Following \cite{shortedTai2015}, we assume that the forget gate $f_{vj}$ depends only on hidden state $h_{vj}$ rather than on the whole context. The term $u_v$ is the value that we use to update the hidden state in a classic Tree-RNN. 
    
    %The Tree-LSTM approximates the unknown state transition function $\hat{f}$ through different neural components computing $i_v$, $o_v$, $u_v$ and $f_{vj}$; these values are then combined together to obtain $c_v$ and $h_v$. The terms $i_v$, $o_v$, $u_v$ are the only ones depending on the context: these three neural elements aggregate the contextual information using a sum-weighted function, as it happens in the state transition function defined in \eqref{eq:sumRNN}. For this reason, in the following, we will refer to this baseline model as Sum-LSTM.
    The values $i_v$, $o_v$, $u_v$ and $f_{vj}$ are computed using a different neural components; among them, only  $i_v$, $o_v$ and $u_v$ depend on the context. Observing Equation \ref{eq:sumLSTM}, it is clear that these three neural elements aggregate the contextual information using a sum-weighted function, as it happens in the state transition function defined in \eqref{eq:sumRNN}. For this reason, in the following, we will refer to this baseline model as Sum-LSTM.
    
    In Section \ref{sec:tensorLSTM}, we generalise the definition of Tree-LSTM using tensor-based aggregation functions to compute the quantities $i_v$, $o_v$, $u_v$. In Section \ref{sec:hosvdLSTM}, we propose a new tensor-based Tree-LSTM model leveraging the HOSVD approximation discussed above. 
    
    \subsection{Full Tensor Tree-LSTM} \label{sec:tensorLSTM}
    The Full Tensor Tree-LSTM (Full-LSTM) is defined by extending the computation of the input gate $i_v$, the output gate $o_v$ and the update value $u_v$ using the general recursive state transition function introduced in \eqref{eq:tensorRNN}, that is
    \begin{equation}
        \begin{split}
            i_v &= \sigma(f^\Tensor{I}(x_v,h_{v1},\dots,h_{v_L})),\\
            o_v &= \sigma(f^\Tensor{O}(x_v,h_{v1},\dots,h_{v_L})),\\
            u_v &= \sigma(f^\Tensor{U}(x_v,h_{v1},\dots,h_{v_L})),
        \end{split}
        \label{eq:full_LSTM}
    \end{equation}
    where $f^\Tensor{I}$, $f^\Tensor{O}$ and $f^\Tensor{U}$ are the multi-affine maps used to compute $i_v$, $o_v$ and $u_v$, respectively. The remainder of the computations needed to complete the forward pass of the LSTM unit can be performed as in \eqref{eq:sumLSTM}. All Full-LSTM parameters are learned using backpropagation. The update rules can be derived easily from the forward computation in \eqref{eq:full_LSTM} using automatic differentiation.
    
    \subsection{HOSVD Tree-LSTM} \label{sec:hosvdLSTM}
    Starting from the tensor-extended formulation of the Full-LSTM, we are now allowed to introduce novel TreeLSTM models leveraging richer and, yet, computationally feasible alternatives to the sum-weighted state transition function. Our proposed HOSVD Tree-LSTM (Hosvd-LSTM) uses the HOSVD approximated state transition function in \eqref{eq:hosvdRNN} to compute the input gate $i_v$, the output gate $o_v$ and the update value $u_v$:
    \begin{equation}
        \begin{split}
            i_v &= \sigma(q^i(g^{\Tensor{G^i}}(w^i(x),u^i_1(h_{1}),\dots,u^i_L(h_{L})))),\\
            o_v &= \sigma(q^o(g^{\Tensor{G^o}}(w^o(x),u^o_1(h_{1}),\dots,u^o_L(h_{L})))),\\
            u_v &= \sigma(q^u(g^{\Tensor{G^u}}(w^u(x),u^u_1(h_{1}),\dots,u^u_L(h_{L})))),
        \end{split}
        \label{eq:HOSVD_LSTM}
    \end{equation}
    where each gate has its own approximation parameters. The superscripts $i$, $o$ and $u$ indicate that a parameter is used to compute $i_v$, $o_v$ and $u_v$ respectively. As before, the remainder of the forward-pass computations can be performed as in \eqref{eq:sumLSTM}. The Hosvd-LSTM update rules can be obtained by backpropagation as for the full tensor case.

\section{Experimental Results}\label{sec:results}
 
    To assess the advantages that we obtain by breaking the relationship between the hidden state size and the parameter space size, we test the Tree-LSTM models introduced in Section \ref{sec:tree-LSTM} on two classification problems.

    \subsection{Experimental Setting}\label{sec:exp_settings}
    In both tasks, tree data are obtained building the syntax tree of input strings according to a specific grammar. In these syntax trees, each label on internal nodes contains an operation that is applied to its child nodes. Hence, we allow models to use different parameters for each operator. It is worth to point out that this parameterisation is consistent with the tensor state transition function. In fact, if we use one-hot encoding to represent operators on internal input labels, the state transition tensor can be sliced obtaining one smaller tensor for each operator; hence, there are no parameters shared between different operators.
    
    To make a fair comparison between different models, we use the label-dependent parametrisation also for the Sum-LSTM: in each internal node, we use the input label (i.e. the operator) to select a different LSTM cell. Also, we do not use word embedding to represent visible labels. In fact, a word embedding layer can map the input label in a space where the aggregation is easier/harder making the comparison unfair.
    
    All the models are trained using AdaDelta algorithm \cite{zeiler2012adadelta} and therefore no learning rate is set. All the weights are initialised using Kaimining normal function \cite{shortedHe2015}. We have implemented all the models using PyTorch \cite{paszke2017automatic} and Deep Graph Library \cite{wang2019dgl}. The code can be found here\footnote{\url{https://github.com/danielecastellana22/tensor-tree-nn}}.

    All the reported results are averaged over three executions, to account for randomisation effects due to initialisation. Accuracy is used to assess model performance in both tasks. Also, we report the number of parameters used to represent a single aggregation function in each model.
    
    %Moreover, we do no to compare our results with ones obtained in \cite{bowman2015} and \cite{shortedNangia2018ListOps:Learning} due to the different experimental setting: they use a different tree representation of the input sentences and they use an input embedding layer.

    \subsection{Relations between Logical Sentences } \label{sec:LRT}
    The goal of the Logical Relations Trees (LRT) task is to determine what kind of logical consequence relation holds between two logical sentences. Each element in the dataset consists of a pair of logical sentences and a relation which exists among them. See \cite{bowman2015} for more details on how the dataset is created.
   
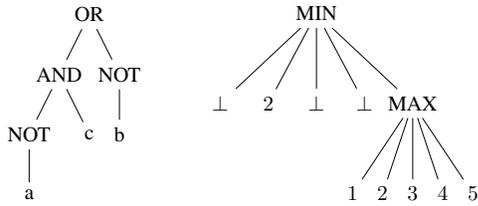
\begin{figure}[tb]
        \centering
            \scalebox{0.8}{
             \begin{tikzpicture}
                    \node (a) at (0,0) {\begin{tikzpicture}[edge from parent/.style={draw,-}, every node/.style = {minimum size = 0.2cm}, level 1/.style={sibling distance=1cm, level distance=1cm}]
                %\footnotesize
                \node {OR}
                child {node {AND}
                    child {node {NOT} 
                        child {node {a}}
                    }
                    child {node {c}}
                }
                child {node {NOT}
                    child {node {b}}
                };
                \end{tikzpicture}};
                \node (b) at (4.5,0){
                    \begin{tikzpicture}[edge from parent/.style={draw,-}]
        	        %\footnotesize
                    \node {MIN} [sibling distance=0.8cm]
                    child {node {$\bot$}}
                    child {node {$2$}}
                    child {node {$\bot$}}
                    child {node (aa) {$\bot$}}
                    child {node {MAX} [sibling distance=0.5cm]
                        child {node {$1$}}
                        child {node {$2$}}
                        child {node {$3$}}
                        child {node {$4$}}
                        child {node {$5$}}
                    };
                    \end{tikzpicture}
                };
                \end{tikzpicture}
            }
        \caption{Example of syntax tree for the LRT (on the left) and ListOps (on the right) dataset.}
        \label{fig:data_example}
    \end{figure}

    Each logical sentence is represented with its syntax tree generated according to the grammar:
    \begin{equation}
        \text{S}:= \,\text{S or S} \mid \text{S and S} \mid \text{not S} \mid \{a,b,c,d,e,f\},
    \end{equation}
    where $\{a,b,c,d,e,f\}$ are boolean variables and \emph{or}, \emph{and} , \emph{not} the common logical operators. In Figure \ref{fig:data_example} we show an example of syntax tree. Due to the arity of the logical operator, the dataset contains only binary trees. Also, the boolean variable are represented using one-hot encoding.
    
    The dataset is already divided into training and test set containing respectively 80\% and 20\% of the data \cite{bowman2015}. We further sample 10\% of the training set in order to build a validation set. During the training we use only trees which contain at most 4 logical operator \cite{bowman2015}, obtaining a training set of around 60k trees. Validation and test set contains around 20k and 36k tree pairs of all size respectively.
        
    The architecture used to solve this task follows the one described in \cite{bowman2015}. We use a single \emph{encoder} model to compress each tree in a fixed size representation (i.e. the root hidden state) and a \emph{classifier} model which predicts one of the seven relations starting from the encoder outputs.
    
    In our experiments, we fix the classifier architecture. Following \cite{bowman2015}, the two trees representation are combined using a tensor and then a non-linear function is applied to the combination result. The non-linear function used is the leaky rectified linear function $\sigma(x) = \text{max}(x, 0) + 0.01 \text{min}(x, 0)$. Finally, a softmax is used to predict the relation. The negative log-likelihood of the correct label with L2 regularisation is the objective function.
    
    %For each Tree-LSTM model used as encoder, we try different configurations. In the Full-LSTM and Sum-LSTM we vary the $c \in \{3,5,10,20,50,100\}$. In the Hosvd-LSTM, we discard the lowest values in order to test more values of the rank; hence, $c \in \{10,20,50,100\}$. For each hidden state size $c$, we try three different values of the rank $r$: $c=10$ and $r \in \{3,5,7\}$, $c=20$ and  $r \in \{5,10,15\}$, $c=50$ and $r \in \{10,20,30\}$,  $c=100$ and  $r \in \{20,50,70\}$.  All the other hyper-parameters are set to the same value for all models: the L2 regularisation weight is set to $0.01$ and the batch size is set to $25$.
    For each Tree-LSTM model used as encoder, we try different configurations. In the Full-LSTM and Sum-LSTM we vary the hidden state size $c \in \{3,5,10,20,50,100\}$. In the Hosvd-LSTM, we try three different values of the rank $r$ for each hidden state size $c$. Hence, we try the following configuration: $c=10$ and $r \in \{3,5,7\}$, $c=20$ and  $r \in \{5,10,15\}$, $c=50$ and $r \in \{10,20,30\}$,  $c=100$ and  $r \in \{20,50,70\}$.  All the other hyper-parameters are set to the same value for all models: the L2 regularisation weight is set to $0.01$ and the batch size is set to $25$.
    
    \subsubsection{Results}
    
       \begin{table*}[ht]
        \centering
        \caption{Validation accuracy and test accuracy on LRT dataset. All scores are averaged over 3 runs (std in brackets).\hspace{\textwidth}Best results are in bold.}
        \footnotesize
        \begin{tabular}{c|c||c|c||c|c||c|c|}
            %\cline{3-8}
            \multicolumn{1}{c}{} &\multicolumn{1}{c||}{} & \multicolumn{2}{c||}{Full-LSTM}  & \multicolumn{2}{c||}{Sum-LSTM}& \multicolumn{2}{c|}{Hosvd-LSTM} \\
            \cline{2-8}
            &$c$ & N par. & Acc. & N par. & Acc. & N par. & Acc.\\
            \hline
            \hline
            \multirow{6}{*}{\rotatebox[origin=c]{90}{Validation}} & $3$ & $48$ & $57.52$ ($0.17$) & $18$ & $57.16$ ($0.25$) & - & - \\
            & $5$ & $180$ & $58.03$ ($0.28$) & $50$ & $57.66$ ($0.12$) & - & - \\
            & $10$ & $1210$ & $84.00$ ($3.23$) & $200$ & $64.79$ ($7.89$) & $588$ & $76.91$ ($4.30$) \\
            & $20$ & $8820$ & $88.77$ ($0.12$) & $800$ & $87.21$ ($1.13$) & $4440$ & $86.99$ ($0.46$) \\
            & $50$ & $130050$ & $91.72$ ($0.78$) & $5000$ & $90.92$ ($0.21$) & $31830$ & $90.29$ ($0.32$) \\
            & $100$ & $1020100$ & $\mathbf{91.86}$ ($0.24$) & $20000$ & $\mathbf{91.33}$ ($0.51$) & $12820$ & $\mathbf{91.11}$ ($0.76$) \\
            \hline
            \hline
            \multirow{2}{*}{\rotatebox[origin=c]{90}{Test}} &  \multirow{2}{*}{$100$} &  \multirow{2}{*}{$1020100$} &  \multirow{2}{*}{$\mathbf{91.46}$ $(0.10)$} &  \multirow{2}{*}{$20000$} &  \multirow{2}{*}{$91.03$ $(0.60)$} &  \multirow{2}{*}{$12820$} &  \multirow{2}{*}{$90.61$ $(0.89)$)} \\
            & & & & & & &\\
            \hline
            \multicolumn{8}{l}{\textbf{Note:} for the Hosvd-LSTM, we report the results
            obtained using the best rank.}
        \end{tabular}
        \label{tab:lrt_acc}
    \end{table*}
    
    \begin{figure}[tb]
        \centering
        \includegraphics[width=\columnwidth]{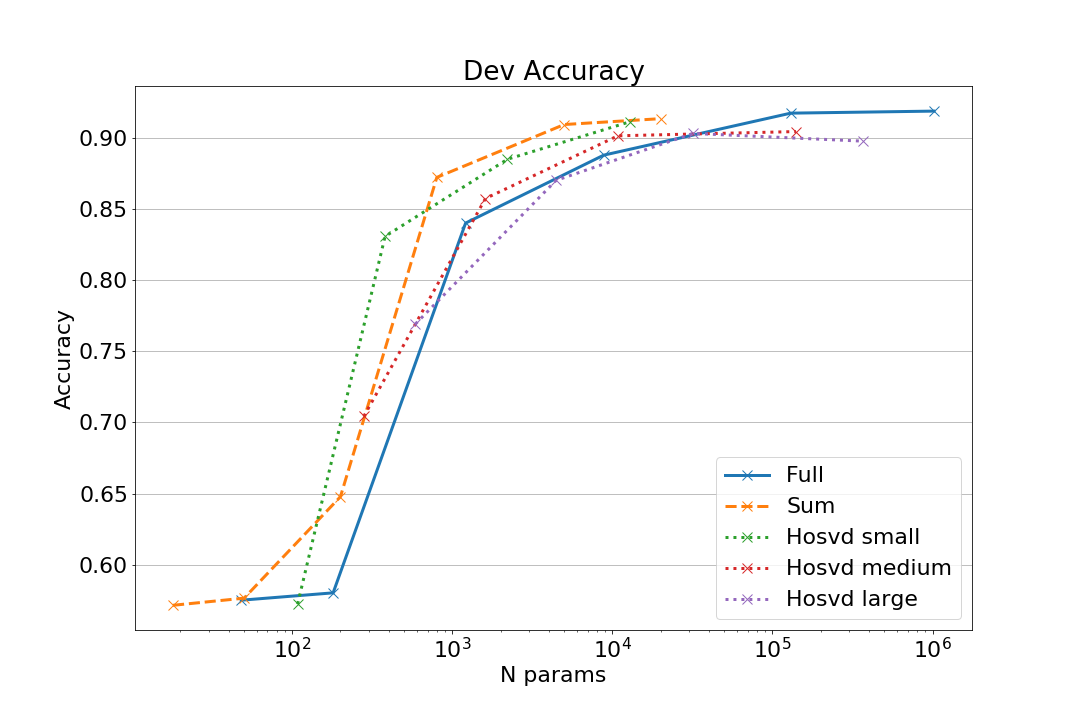}
        \caption{Validation accuracy against number of parameters for each configuration of Full-LSTM, Sum-LSTM, Hosvd-LSTM on LRT dataset. For the Hosvd-LSTM, we plot one line for each rank values: \textit{small}, \textit{medium} and \textit{large} indicates the smallest, the central and the largest rank values respectively.}
        \label{fig:lrt_acc}
    \end{figure}
    
    The results obtained on the validation set using different configurations of Full-LSTM, Sum-LSTM and Hosvd-LSTM are reported in Table \ref{tab:lrt_acc}. Also, the same results are plotted in Figure \ref{fig:lrt_acc} against the number of parameters. For the Hosvd-LSTM, we report only the best result obtained for each value of $c$; all the results obtained by other Hosvd-LSTM configurations are reported in the supplementary material.
        
    The performances obtained by all models are similar for all configurations except the one which set $c=10$. In that case, using the Full-LSTM we obtain an accuracy of $84.00\%$; with the same configuration, the Sum-LSTM reaches an accuracy of $64.79\%$ while the Hosvd-LSTM reaches $76.91\%$. This suggests that the model with $c<10$ are not able to solve the task because the hidden state size is too small to encode all the necessary structural information. Setting $c>10$ yields to a data representation which is rich enough to solve the task using a simple combination of them such as the sum. Instead, setting $c=10$, we observe the advantage of leveraging a more general aggregation function; the expressivity of such function allows us to achieve satisfactory results using a small hidden state size. The price to pay is, of course, a larger number of parameters. Also, it is interesting to compare the best configurations for each model. All models reach similar performances (see Table \ref{tab:lrt_acc}), but the number of parameters used by the best Hosvd-LSTM is half of the parameters used by the best Sum-LSTM and a hundredth of the parameters used by the best Full-LSTM.
     
    In Table \ref{tab:lrt_acc}, we also report the accuracy on the test set obtained by the best models configurations. The Full-LSTM and Sum-LSTM best configuration is $c=100$; the Hosvd-LSTM best configuration is $c=100$ and $r=20$.
    
    Note that our results cannot be compared with the ones obtained in \cite{bowman2015} due to the different experimental settings, i.e. the different parsing strategies used to build trees and the encoding of input labels.
    
    \iffalse
    \begin{table}
        \centering
        \caption{Test accuracy on LRT dataset}
        \footnotesize
            \begin{tabular}{c|c}
                Model & Test Accuracy\\
                \hline
                \hline
                Full & $\mathbf{91.46}$ $(0.10)$ \\
                Sum & $91.03$ $(0.60)$ \\
                Hosvd & $90.61$ $(0.89)$
            \end{tabular}
        %\caption{Test accuracy on LRT dataset. All scores are averaged over 3 runs (standard deviations in parentheses). Best results are in bold. }
        \label{tab:lrt_test_acc}
    \end{table}
    \fi

    \subsection{Operation on List} \label{sec:ListOps}
    The goal of the ListOps task is to predict the solution of a sequence of summary operations on lists of single-digit integers, written in prefix notation. Each element in the dataset consists of a sequence of operations and its solution (which is also a single-digit integer). See \cite{shortedNangia2018} for more details on the dataset generation. Each sequence is represented as a tree using its syntax tree built according to the following grammar:
    \begin{equation}
        \begin{split}
            \text{S}:= \, &\text{max[S,S,S,S,S]} \mid \text{min[S,S,S,S,S]} \mid \text{sum-m[S,S,S,S,S]} \mid\\
            &\text{med[S,S,S,S,S]} \mid \{0,\dots,9\} \mid \bot,
        \end{split}
    \end{equation}
    where the bottom symbol $\bot$ represents a missing operand. We show an example of syntax tree in Figure \ref{fig:data_example}. Each operation has at most five inputs; hence, trees in this dataset have a maximum output degree of five (i.e. $L=5$). Also, the input digit $k$ is represented using a vector $x$ of size  $10$ which has the first $k+1$ entries equals to 1 and the other equals to 0 (e.g. if $k=2$, $x=[1,1,1,0,0,0,0,0,0,0]$).
    
    \iffalse
    \begin{figure}[tb]
        \centering
        \scalebox{0.8}{
            \begin{tikzpicture}[edge from parent/.style={draw,-}, every node/.style = {minimum size = 0cm},node distance = 0.05cm, level1/.style={sibling distance=0.01cm}]
            \node {MIN}
            child {node {$\bot$}}
            child {node {$2$}}
            child {node {$\bot$}}
            child {node {MAX}
                child {node {$1$}}
                child {node {$2$}}
                child {node {$3$}}
                child {node {$4$}}
                child {node {$5$}}
            }
            child {node {$\bot$}};
            \end{tikzpicture}}
        \caption{Syntax tree of the sequence $\text{MIN}[2, \text{MAX}[1,2,3,4,5]]$.}
        \label{fig:ListOps_example}
    \end{figure}
    \fi
    
    The dataset is already divided into training and test splits containing respectively 90\% and 10\% of the data \cite{shortedNangia2018}. We further sample 9\% of the training set in order to build a validation set. Hence, we obtain a training set which contains around 80k trees; validation and test set contains around 20k and 36k trees respectively.
    
    The architecture used to tackle this task is straightforward: we use a Tree-LSTM model to encode trees in a fixed size representation (i.e. the root hidden state) and then a classifier to predict the correct result starting from the tree representation. In our experiments, the classifier is fixed and it is implemented as a two-layer neural network with $20$ hidden units for each layer. The cost function is the negative log-likelihood with respect to the ground-truth label with an L2 regularisation term. 
    
    For each Tree-LSTM model used as encoder, we test multiple configurations. In the Full-LSTM, we vary the hidden state size $c \in \{3,5,7\}$; in the Sum-LSTM, we vary $c \in \{25,88,214\}$; in the Hosvd-LSTM, we vary $c \in \{10,20,50\}$ and $r \in \{3,5,7\}$. The values of $c$ and $r$ are chosen to have models with the same number of parameters. All the other hyper-parameters are set to the same value for all models: the L2 regularisation weight is set to $0.01$ and the batch size is set to $25$.
    
    \subsection{Results}
\begin{figure}[!th]
        \centering
        \includegraphics[width=\columnwidth]{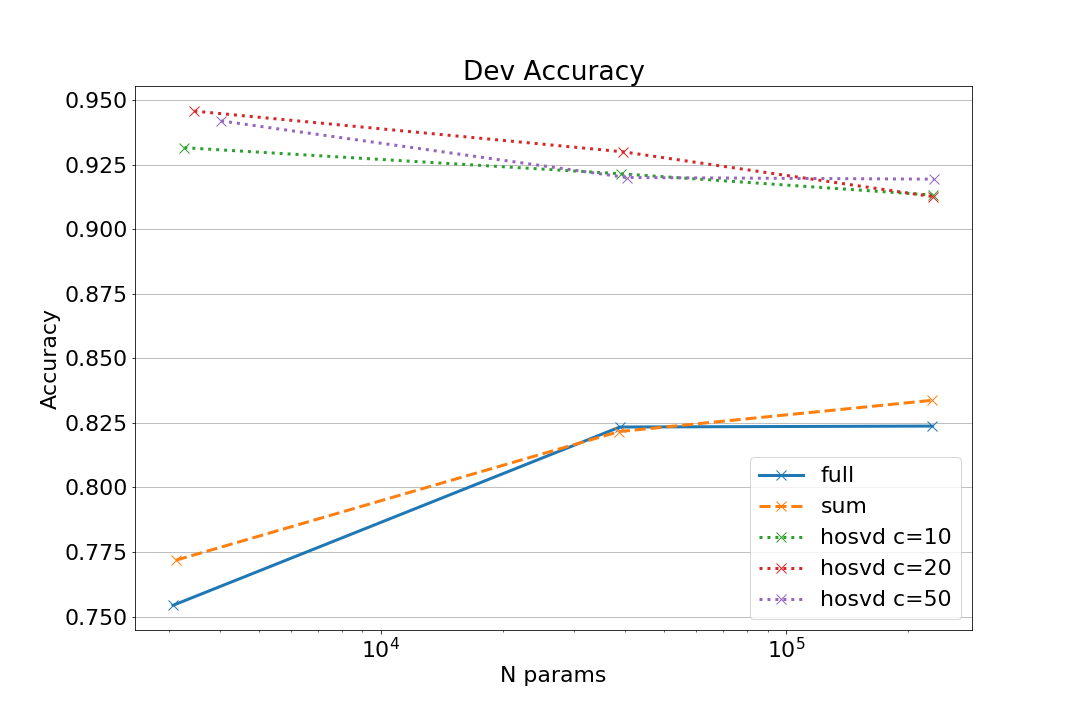}
        \caption{Validation accuracy against number of parameters for each configuration of Full-LSTM, Sum-LSTM, Hosvd-LSTM on ListOps dataset. For the Hosvd-LSTM, we plot one line for each hidden state size $c$.}
        \label{fig:ops_acc}
    \end{figure}

    The results obtained on the validation set using different configurations of Full-LSTM, Sum-LSTM and Hosvd-LSTM are reported in Table \ref{tab:ops_acc}. Also, the same results are plotted in Figure \ref{fig:ops_acc} against the number of parameters. For the Hosvd-LSTM, we report only the best result obtained for each value of $c$; all the results obtained by other Hosvd-LSTM configurations are reported in the supplementary material. These results highlight the superior performance of Hosvd-LSTM compared to the other models. In particular, all Hosvd-LSTM configurations reach an accuracy that is higher than all configurations of the Sum-LSTM and Full-LSTM. This result becomes even more interesting if we compare the number of parameters used by the best Hosvd-LSTM model: it uses only 3k parameters, while Full-LSTM and Sum-LSTM are not able to reach the same performance even with more than 1 million parameters. The accuracy of Hosvd-LSTM decreases with increasing the number of parameters, an aspect which needs further investigations.
    
    We argue that the Full-LSTM is not able to obtain satisfactory performance due to the small hidden state size. In fact, we vary $c \in \{3,5,7\}$; larger values of $c$ would lead to a huge number of model parameter due to the maximum output degree of this dataset (i.e. $L=5$). Nevertheless, also the Sum-LSTM performs poorly on this dataset. We argue that this is due to the aggregation function which is too simple to implement operations in the dataset. Therefore, the Hosvd-LSTM outperforms the other two models thanks to the new approximation which allows combining a powerful aggregation function with a large hidden state size.
    
    In Table \ref{tab:ops_acc}, we also report the accuracy on the test set obtained by the best model configurations. The Full-LSTM best configuration is $c=7$; the Sum-LSTM best configuration is $c=214$; the Hosvd-LSTM best configuration is $c=20$ and $r=3$.
    
    %Our results cannot be compared with the ones obtained in \cite{shortedNangia2018ListOps:Learning} due to the different experimental settings, i.e. the different parsing strategy used to build trees and the encoding of input labels.
    
    \iffalse
    \begin{table*}
        \centering
        \footnotesize
        \begin{tabular}{||c|c|c||c|c|c||c|c|c||}
            \multicolumn{3}{||c||}{Full-LSTM} & \multicolumn{3}{c||}{Sum-LSTM} & \multicolumn{3}{c||}{Hosvd-LSTM}\\
            \hline
            $c$ & N par. & Acc. & $c$ & N par. & Acc. & $c$ & N par. & Acc. \\
            \hline
            \hline
            $3$ & 3072 & $75.30$ ($2.74$) &  $25$ & 3125 & $77.88$ ($0.85$) & 10 & $3222$ & $93.15$ ($0.61$)\\
            $5$ & 38880 & $82.76$ ($0.36$) & $88$ & 38720 & $82.80$ ($1.08$) & 20 & $3372$ & $\mathbf{94.57}$ ($0.70$) \\
            $7$ & 229376 & $82.48$ ($0.59$)& $214$ & 228980 & $84.51$ ($0.30$) & 50 & $3822$ &  $94.19$ ($0.30$) \\
            $10$ & 1610510 & $\mathbf{82.99}$ ($0.69$) & $568$ & 1613120 & $\mathbf{84.99}$ ($0.23$) & 100 & $4572$ & $92.84$ ($0.27$)\\
        \end{tabular}
        \caption{Validation accuracy and number of parameters for each Full-LSTM, Sum-LSTM and Hosvd-LSTM configuration on the ListOps dataset. For the Hosvd-LSTM, we report results obtained using the best rank. All scores are averaged over 3 runs (standard deviations in parentheses). Best results are in bold.}
        \label{tab:ops_acc}
    \end{table*}
    \fi
    
    \begin{table*}[t]
        \centering
        \caption{Validation accuracy and test accuracy on ListOps dataset. All scores are averaged over 3 runs (std in brackets).\hspace{\textwidth}Best results are in bold.}
        \label{tab:ops_acc}
        \footnotesize
        \begin{tabular}{c||c|c|c||c|c|c||c|c|c|}
            & \multicolumn{3}{c||}{Full-LSTM} & \multicolumn{3}{c||}{Sum-LSTM} & \multicolumn{3}{c|}{Hosvd-LSTM}\\
            \cline{2-10}
            & $c$ & N par. & Acc. & $c$ & N par. & Acc. & $c$ & N par. & Acc. \\
            \hline
            \hline
            \multirow{3}{*}{\rotatebox[origin=c]{90}{Val.}} & $3$ & 3072 & $75.44$ ($1.18$) & $25$ & 3125 & $77.19$ ($0.81$) & 10 & $3222$ & $93.15$ ($0.61$)\\
            & $5$ & 38880 & $82.34$ ($0.61$) & $88$ & 38720 & $82.16$ ($0.98$) & 20 & $3372$ & $\mathbf{94.57}$ ($0.70$) \\
            & $7$ & 229376 & $\mathbf{82.38}$ ($0.82$) & $214$ & 228980 & $\mathbf{83.37}$ ($1.18$) & 50 & $3822$ &  $94.19$ ($0.30$) \\
            \hline
            \hline
            \multirow{2}{*}{\rotatebox[origin=c]{90}{Test}} &  \multirow{2}{*}{$7$} &  \multirow{2}{*}{$229376$} &  \multirow{2}{*}{$82.02$ $(0.65)$} &  \multirow{2}{*}{$214$} &  \multirow{2}{*}{$228980$} &  \multirow{2}{*}{$83.01$ $(1.06)$} & \multirow{2}{*}{$20$} &  \multirow{2}{*}{$3372$} & \multirow{2}{*}{$\mathbf{94.26}$ $(0.48)$} \\
            & & & & & & & & &\\
            \hline
            \multicolumn{10}{l}{\textbf{Note:} for the Hosvd-LSTM, we report the results
            obtained using the best rank.}
        \end{tabular}
        %\caption{Validation accuracy and number of parameters for each Full-LSTM, Sum-LSTM and Hosvd-LSTM configuration on the ListOps dataset. For the Hosvd-LSTM, we report results obtained using the best rank. All scores are averaged over 3 runs (standard deviations in parentheses). Best results are in bold.}
    \end{table*}
    
\iffalse
    \begin{table}
        \centering
        \caption{Test accuracy on ListOps dataset}
        \footnotesize
            \begin{tabular}{c|c}
                Model & Test Accuracy\\
                \hline
                \hline
                Full & $82.02$ $(0.65)$ \\
                Sum & $83.01$ $(1.06)$ \\
                Hosvd & $\mathbf{94.26}$ $(0.48)$\\
                %\multicolumn{2}{l}{} \\
                %\multicolumn{2}{l}{All scores are averaged over 3 runs.} \\
                %\multicolumn{2}{l}{Best results are in bold.}
            \end{tabular}

        \label{tab:ops_test_acc}
    \end{table} 
\fi
\section{Conclusion}\label{sec:conclusion}
In this paper, we have introduced a general framework to model context aggregation in tree-structured data based on a tensor formulation. In particular, we have shown that a Bottom-Up parsing direction of the input data leads to the definition of a tensor-based state transition function. The size of such tensor increases exponentially with respect to the maximum out-degree of input trees, imposing strong limitations on the choice of the hidden state size.

The central contribution of this paper is an approximation of such tensor which leverages the HOSVD tensor decomposition. This approximation allows breaking the exponential relation between the number of model parameters and the hidden state size without losing a tensor-based aggregation function. The results of our experiments show the advantage of our approximation, especially when the maximum out-degree of the tree increases as, for instance, in the ListOps dataset.

Such results pave the way to the definition of a  tensor-based framework for structured data processing, stimulating the development of new adaptive tree models which leverage tensor factorisation. To this end, the next step would be to explore the applicability of other tensor factorisations, such as the canonical decomposition and the tensor train decomposition, and to study which are the bias introduced by each of them.

Ultimately, we would like to study how to include in our theoretical framework other tree model architectural biases. For example, we believe that positional stationarity can be implemented in our framework by adding a symmetry constraint on the state transition tensor.

\section*{Acknowledgment}
The work is supported by project MIUR-SIR 2014 LIST-IT (grant n. RBSI14STDE).

\bibliography{biblio, others, related_work}

\end{document}